\documentclass{article}

\usepackage[preprint]{neurips_2022}
\usepackage[utf8]{inputenc} 
\usepackage[T1]{fontenc}   
\usepackage[colorlinks=true,linkcolor=blue,urlcolor=blue,citecolor=blue,anchorcolor=blue]{hyperref}
\usepackage{url}       
\usepackage{booktabs}
\usepackage{amsfonts,amsmath,amssymb} 
\usepackage{nicefrac}
\usepackage{microtype}
\usepackage{xcolor}
\usepackage[pdftex]{graphicx}
\usepackage{enumitem}
\usepackage{multirow}
\usepackage{makecell}
\usepackage{tablefootnote}
\usepackage{caption}
\usepackage{wrapfig}
\usepackage{subfigure}
\usepackage{comment}

\usepackage{titlesec}
\titlespacing\section{2pt}{2pt plus 2pt minus 2pt}{0pt plus 1pt minus 1pt}
\titlespacing\subsection{2pt}{1pt plus 2pt minus 1pt}{0pt plus 1pt minus 1pt}
\titlespacing\subsubsection{2pt}{1pt plus 2pt minus 1pt}{0pt plus 1pt minus 1pt}

\title{Self-Supervised Pretraining\\ for Differentially Private Learning}

\author{
  Arash Asadian\\
  Indiana University\\
  \texttt{aasadian@iu.edu} \\
  \And
	Evan Weidner\\
  Indiana University\\
  \texttt{eweidne@iu.edu} \\
	\And
  Lei Jiang \\
  Indiana University\\
  \texttt{jiang60@iu.edu} \\
}

\begin{document}

\maketitle

\begin{abstract}
We demonstrate self-supervised pretraining (SSP) is a scalable solution to deep learning with differential privacy (DP) regardless of the size of available public datasets in image classification. When facing the lack of public datasets, we show the features generated by SSP on only one single image enable a private classifier to obtain a much better utility than the non-learned handcrafted features under the same privacy budget. When a moderate or large size public dataset is available, the features produced by SSP greatly outperform the features trained with labels on various complex private datasets under the same private budget. We also compared multiple DP-enabled training frameworks to train a private classifier on the features generated by SSP. Finally, we report a non-trivial utility 25.3\% of a private ImageNet-1K dataset when $\epsilon=3$. Our source code can be found at \url{https://github.com/UnchartedRLab/SSP}.
\end{abstract}

\section{Introduction}
\label{s:intro}

Machine learning (ML) has been applied ubiquitously in the analysis of sensitive data such as medical images~\citet{Tajbakhsh:TMI2016}, financial records~\citet{Fischer:EJOR2018}, or social media channels~\citet{Agrawal:ECIR2018}. Many attacks~\citet{Shokri:SP2017,Carlini:USENIX2021} are developed to successfully extract meaningful training data out of standard ML models. According to recent regulations, e.g., GDPR and CCPA, ML models have to protect sensitive training data. \textit{Differential privacy} (DP)~\cite{Chaudhuri:JMLR2011,Bu:HDSR2020,Abadi:CCS2016} has emerged as an effective framework to train models resilient to private training data leakage.

Unfortunately, training models with strong DP guarantees significantly hurts the model utility (i.e., accuracy)~\citet{Papernot:ICLR2018,Abadi:CCS2016}. Although non-learned handcrafted features (i.e., ScatterNet~\citet{Oyallon:CVPR2015,Oyallon:TPAMI2018}) make a private linear model~\citet{Tramer:ICLR2021} achieve the state-of-the-art (SOTA) utility of $<70\%$ under the privacy budget of $( \epsilon \leq 3, \delta=10^{-5})$ on a private CIFAR-10 dataset, it is difficult to learn better features in the DP domain, since the clipped and perturbed gradients during DP training provide only a noisy estimate of the update direction. In contrast, it is straightforward that pretrained features learned from large public labeled~\citet{Lou:CVPR2021} datasets can greatly mitigate the utility gap between private and non-private models. However, sometimes there is no available public dataset for training a feature extractor due to legal causes or ethical issues~\citet{Flanders:RADIO2009}.

%Among all DP-enabled training frameworks, DPDFA is more effective when learning on pretained features. 
%When further trained by a small number of (i.e., 1K) unlabeled public images, the feature extractor greatly improves the private model utility over the handcrafted features, as shown in Table~\ref{t:privacy_cifar10_motivation}. 
%Further training the feature extractor with 1K unlabeled public images greatly improves the private classifier utility over the non-learned handcrafted features.
%DPSGLD greatly reduces the privacy overhead when training with high-quality pretained features.
%using feature extractors built upon various network architectures, e.g., CNNs and ViTs~\citet{Dosovitskiy:ICLR2021}, and trained on large public datasets. Particularly, we show a private classifier achieves the utility of 50\% under the privacy budget of $( \epsilon \leq 3, \delta=10^{-5})$ on a private ImageNet-1k dataset~\citet{Krizhevsky:NIPS2012} using the pretrained features learned on .

In this paper, we aim to demonstrate that \textit{self-supervised pretraining} (SSP) \textit{is a scalable solution to improving the utility of deep learning with DP regardless of the size of available public datasets} in image classification. Any updates on the learnable parameters of a differentially private model increase privacy overhead. It is easier to achieve both high utility and small privacy loss via the features generated by a well-trained feature extractor that can fully take advantage of SOTA network architectures and public datasets. Even when no large public dataset is available, we show a feature extractor built upon data-efficient HarmonicNet~\citet{Ulicny:BMVC2019} and trained by self-supervised SimCLRv2~\cite{Chen:NIPS2020} on \textit{only one single image}~\citet{Asano:ICLR2020} can make a private linear classifier obtain much better utility than the non-learned handcrafted features~\citet{Tramer:ICLR2021} under the same privacy budget. With a larger public dataset, the features generated by SSP substantially outperform the features trained with labels on various complex private datasets, as shown in Table~\ref{t:privacy_cifar10_motivation}. To better explore the trade-off between utility and privacy, we compared SOTA DP-enabled training frameworks, i.e., DP stochastic gradient descent (DPSGD)~\citet{Abadi:CCS2016}, DP direct feedback alignment (DPDFA)~\citet{Ohana:NIPS2021,Lee:CORR2020}, DP stochastic gradient Langevin dynamics (DPSGLD)~\citet{Bu:ICMLTPDPW2021}, and Private Aggregation of Teacher Ensembles (PATE)~\citet{Papernot:ICLR2018}, when training a private classifier on the features produced by SSP. Private datasets having different learning distances from the public dataset favor different training frameworks under different privacy budgets. Our contributions are summarized as the follows.
\begin{itemize}[leftmargin=*, topsep=0pt, partopsep=0pt, itemsep=0pt]

\item When facing the lack of public datasets, we adopt HarmonicNet as the backbone of SimCLRv2 to learn only one image. The features extracted by the HarmonicNet greatly outperform the non-learned handcrafted ScatterNet features on various private datasets by 0.6\% on CIFAR10, 1.4\% on CIFAR100, 6.7\% on CropDiseases, 49.7\% on EuroSAT, and 3.5\% on ISIC2018, when $\epsilon=2$.

\item When there is a moderate or large size public dataset, the features produced by SSP improve the utility of these private complex datasets over the features trained by labels by $0.5\%\sim8.6\%$ under the privacy budget of $\epsilon=2$.

\item We compared SOTA DP-enabled training frameworks, i.e., DPSGD, DPDFA, DPSGLD, and PATE, to train a private classifier on the features produced by SSP. Compared to DPSGD, DPSGLD obtains a better utility on private datasets when $\epsilon\leq 1$. DPDFA achieves a higher utility than DPSGD on private datasets with a smaller learning distance from the public dataset when $\epsilon> 0.5$.

\item Through the features produced by SSP on a large unlabeled public dataset PASS~\citet{Asano:PASS2021}, we report a non-trivial utility 25.3\% of a private ImageNet-1K dataset when $\epsilon=3$.
\end{itemize}

\begin{table}[t!]
\caption{The utility and privacy comparison on a private CIFAR10 dataset. DP-SOTA-1 means \citet{Tramer:ICLR2021}, and DP-SOTA-2 indicates \citet{Lou:CVPR2021}.}
\vspace{+0.1in}
\label{t:privacy_cifar10_motivation}
\centering
\begin{tabular}{cccccc} \toprule
Public Dataset  & Scheme        & $\epsilon$-DP  & Network     & Training    & Accuracy (\%)     \\ \midrule
None            & DP-SOTA-1     & 3              & Scat + CNN  & DPSGD       & 66.1\tablefootnote{Group normalization. The results of batch normalization can be found in Appendix~\ref{s:normalization}.}       \\
1 image         & \textbf{Ours} & 3              & \textbf{HarmRN18}  & DPSGD       & \textbf{66.3}\tablefootnote{Batch normalization trained by the public dataset.}\\ \midrule

labeled CIFAR100   & DP-SOTA-2      & 1.5        & ResNet18    & DPSGD       & 71\footnotemark[1]       \\ 
\textbf{unlabeled} CIFAR100 & \textbf{Ours}  & 1.5        & ResNet18    & DPSGD       & \textbf{75.5}\footnotemark[2]      \\ \midrule
\multirow{2}{*}{unlabeled ImageNet}   & DP-SOTA-1     & 2      & ResNet50    & DPSGD  & 92.7              \\ 
                                      & \textbf{Ours} & 2      & ResNet50    & \textbf{DPDFA}  & \textbf{94.9}                  \\ \bottomrule
\end{tabular}
\vspace{-0.2in}
\end{table}

\section{Background}
\label{s:back}

\subsection{Differentially Private Learning}

\textbf{Differential privacy}. A network model $M:\mathcal{D}\rightarrow \mathcal{R}$ is trained on two datasets $D, D'\in \mathcal{D}$, which differ only by a single data record. For any subset of outputs $R\in\mathcal{R}$, the model satisfies  $(\epsilon, \delta)$-differential privacy (DP)~\citet{Abadi:CCS2016} if
\begin{align*}
\text{\bf Pr}[M(D)\in R] \leq e^{\epsilon} \cdot \text{\bf Pr}[M(D')\in R] + \delta
\end{align*}
In another word, $\epsilon$ bounds the privacy loss on any single sample, and $\delta$ is the probability that this bound does not hold. R\'enyi DP (RDP)~\citet{Mironov:CSF2017} is a generalization of $(\epsilon, \delta)$-DP that uses R\'enyi divergence as a distance metric. More RDP details can be viewed in Appendix~\ref{s:renyi}.

\textbf{DP-enabled training}. DP guarantees can be enforced into a private network model by various DP-enabled training frameworks including DPSGD~\citet{Abadi:CCS2016}, DPSGLD~\citet{Bu:ICMLTPDPW2021}, DPDFA~\citet{Ohana:NIPS2021,Lee:CORR2020}, and PATE~\citet{Papernot:ICLR2018}.
\begin{itemize}[leftmargin=*, topsep=0pt, partopsep=0pt, itemsep=0pt]
\item \textbf{DPSGD}. DPSGD~\cite{Abadi:CCS2016} is the most widely used DP-enabled training framework that satisfies RDP. RDP during DPSGD on a model is enforced by two components: (1) per-sample gradients are clipped at a fixed L2 norm threshold $C$; and (2) Gaussian noise of magnitude $\sigma^2 C^2$ is added to the gradient updates for a noise scale parameter $\sigma$. The privacy cost $\epsilon$ during DPSGD can be measured by Moments Accountant~\citet{Abadi:CCS2016}, which computes the upper bound of $\epsilon$ as a consequence of using different composition theories. The privacy loss rate depends on the hyper-parameters~\citet{Shubhankar:CORR2021} of DPSGD. 

%As a result, the privacy loss rate depends on the DPSGD training hyperparameters including the number of epochs $T$, the learning rate $\eta$, the noise scale $\sigma$, the batch size $|B|$, and the clipping norm $C$.

\item \textbf{DPSGLD}. SGLD~\citet{Welling:ICML2011} is a gradient technique to train Bayesian networks. SGLD makes the weights of a model to converge to a posterior distribution rather than to a point estimate, from which it can sample and characterize the uncertainty of weights. DPSGLD~\citet{Wang:ICML2015} is built to train DP Bayesian networks by adding per-sample clipping. Recent work~\cite{Bu:ICMLTPDPW2021} proves that DPSGLD is equivalent to DPSGD with regularization~\cite{Bu:ICMLTPDPW2021}. The equivalence can be described as
\begin{align*}
\text{DPSGLD}(\eta_{SGLD}=\eta, C_{SGLD}=C)=\text{DPSGD}(\eta_{SGD}=\eta N, \sigma_{SGD}=\frac{|B|}{N\sqrt{\eta}C},C_{SGD}=C),
\end{align*}
where $\eta_X$ is the learning rate of $X$; $C_{X}$ is the clipping norm of $X$; $|B|$ is the batch size; and $N$ is the private dataset size.

\item \textbf{DPDFA}. DFA~\citet{Arild:NIPS2016} emerges as an effective alternative to backpropagation in non-private training. DPDFA~\citet{Ohana:NIPS2021,Lee:CORR2020} enforces DP in DFA by clipping activations and error signal after the feedforward phase, scaling the error transport matrix, and adding Gaussian noise to the update direction.

\item \textbf{PATE}. PATE~\citet{Papernot:ICLR2018} is a framework based on private knowledge aggregation of an ensemble model and knowledge transfer. It trains an ensemble of \textit{teachers} on disjoint subsets of the private dataset. The ensemble's knowledge is then transferred to a \textit{student} model via differentially private aggregation of the teachers’ votes on samples from an unlabeled public dataset. The student model is released as the output of the training.
\end{itemize}

\subsection{Handcrafting and Learning Low-level Features}

\textbf{Handcrafted features}. Recent work~\citet{Tramer:ICLR2021} advocates adopting the wavelet-transform-based ScatterNet~\citet{Oyallon:CVPR2015,Oyallon:TPAMI2018} to extract non-learned handcrafted features (e.g., invariance to small rotations and translations) for differentially private learning. A private linear classifier learned on these handcrafted features exhibits higher utility than an end-to-end CNN under a moderate privacy budget. 

\textbf{Learned features}. In the non-private setting, recent work~\citet{Asano:ICLR2020} demonstrates the features learned from one single image with strong augmentations and without supervision can obtain a non-trivial accuracy on ImageNet. However, the performance of these learned features is not studied in the DP domain. In this paper, we show that the learned features are more effective than the handcrafted ScatterNet features in the DP domain. %Moreover, we demonstrate these features further trained by a small number (1K) of unlabeled images can greatly outperform the ScatterNet features. 

\textbf{Combining handcrafted and learned features}. HarmonicNet~\citet{Ulicny:BMVC2019} is a data-efficient network using a set of non-learned handcrafted filters based on Discrete Cosine Transform (DCT) at multiple frequencies combined by learnable weights. The DCT-based filters have excellent energy compaction properties. It can obtain higher non-private accuracy~\citet{Ulicny:BMVC2019} than ScatterNet, particularly when only small training datasets are available. However, the performance of the features generated by a HarmonicNet architecture is never examined in the DP domain.

\begin{table}[t!]
\centering
\begin{minipage}[b]{0.48\linewidth}
\centering
\captionof{table}{The utility comparison of private models directly trained on a private CIFAR10 under $(\epsilon=3, \delta=10^{-5})$.}
\vspace{+0.1in}
\label{t:privacy_cifar10_noimage}
\begin{tabular}{cccc} \toprule
Public  & \multicolumn{2}{c}{Network}       & \multirow{2}{*}{Acc (\%)} \\
Dataset & \multicolumn{2}{c}{Architecture}  &                           \\ \midrule
\parbox[t]{2mm}{\multirow{5}{*}{\rotatebox[origin=c]{90}{\textbf{none}}}} & \multicolumn{2}{c}{end-to-end CNN}  & 59.2  \\
        & \multicolumn{2}{c}{ScatterNet+linear} & 64.2 \\
				& \multicolumn{2}{c}{\textbf{ScatterNet+CNN}}    & \textbf{66.1} \\
				& \multicolumn{2}{c}{HarmonicNet + linear} & 47.7 \\
				& \multicolumn{2}{c}{HarmonicNet + CNN}    & 50.2 \\ \bottomrule
\end{tabular}
\end{minipage}
\hfill
\begin{minipage}[b]{0.48\linewidth}
\centering
\captionof{table}{The utility comparison of private linear classifiers learned on pretrained features and a private CIFAR10 under $(\epsilon=3, \delta=10^{-5})$.}
\vspace{+0.1in}
\label{t:privacy_cifar10_fewimage}
\begin{tabular}{cccc} \toprule
Public  & \multicolumn{2}{c}{Feature}    & \multirow{2}{*}{Acc (\%)} \\
Dataset & \multicolumn{2}{c}{Extractor}  &                           \\ \midrule
\parbox[t]{2mm}{\multirow{6}{*}{\rotatebox[origin=c]{90}{\textbf{one image}}}} & \parbox[t]{2mm}{\multirow{4}{*}{\rotatebox[origin=c]{90}{ResNet18}}} & ResBlock-1 & 49.7 \\       
        &  & ResBlock-2 & 56.3\\
				&  & ResBlock-3 & 51.4\\
				&  & ResBlock-4 & 50.3\\
				& \multicolumn{2}{c}{ScatRN18}   & 63.1\\
				& \multicolumn{2}{c}{\textbf{HarmRN18}}   & \textbf{66.3} \\\bottomrule
\end{tabular}
\end{minipage}
\vspace{-0.2in}
\end{table}

\section{Self-Supervised Pretraining For Differentially Private Learning}
\label{s:pretain}

In this section, we explain how to build scalable pretrained features for differentially private learning with various sizes of public datasets, i.e., no public dataset, a moderate size (38K$\sim$50K) public dataset, and a large size (>1M) public dataset. Although batch normalization (BN) can be supported by DP in principle, it is difficult to use the same hyper-parameters, e.g., noise multiplier and learning rate, to train BN parameters and the other network parameters in the DP domain. The SOTA DP-enabled training libraries such as Opacus~\cite{Yousefpour:NWPML2021} do not even support BN yet. Therefore, in this paper, we enforce that DP-enabled training directly happening on the private dataset requires the model to use only group normalization, and all pretrained features are generated with the BN parameters trained by only the public dataset. How to fine-tune BN parameters with private datasets is beyond the topic of this paper, and can be answered by~\citet{Lou:CVPR2021}. Results of BN on private datasets can be viewed in Appendix~\ref{s:normalization}. The utility and privacy improvement of the features generated by SSP over non-learned handcrafted features and those trained with labels is not reduced by private BN.

\subsection{No public dataset is available}
\label{s:nolarge}

When no public dataset is available, we compare two ways to train a private model on a private dataset. One way is to train a private model directly on the private dataset using non-learned handcrafted features. The other way is to train a feature extractor by a data-efficient network architecture, self-supervision, and a single image, and then to train a private linear classifier on the private dataset using features produced by the feature extractor. We used only DPSGD for DP training in this section.

\textbf{Training directly on the private dataset}. We show the utility comparison of various schemes directly learning a private CIFAR10 under the privacy budget of $(\epsilon=3, \delta=10^{-5})$ in Table~\ref{t:privacy_cifar10_noimage}. The training schemes include a 5-layer CNN, a linear classifier learned on the ScatterNet handcrafted features, and a 5-layer CNN learned on the ScatterNet features, which are adopted from~\citet{Tramer:ICLR2021}. We also compare a linear classifier and a 5-layer CNN learned on the DCT-based HarmonicNet handcrafted features. The CNN with the ScatterNet features achieves the highest utility, i.e., 66.1\%. We find the HarmonicNet features are vulnerable to noises, i.e., applying noises to the weights combining multiple DCT frequencies significantly degrades the utility of private models.

%These augmentations are representative of invariances that we want to incorporate in the features. Augmentation can be seen as imposing a prior on how we expect the manifold of natural images to look like. When training with very few images, these priors become more important since the model cannot extract them directly from data.

\begin{table}[t!]
\centering
\begin{minipage}[b]{0.48\linewidth}
\centering
\captionof{table}{The CIFAR10 utility comparison of the features produced by feature extractors further fine-tuned by different numbers of public images from CIFAR100 under $(\epsilon=3, \delta=10^{-5})$.}
\vspace{+0.1in}
\label{t:privacy_cifar10_tuning}
\begin{tabular}{ccccc} \toprule
\multirowcell{3}{Pretrained\\ Feature\\ Extractor} & \multicolumn{4}{c}{Acc (\%) $(\epsilon=3, \delta=10^{-5})$}  \\ \cmidrule(r){2-5}
               & \multicolumn{4}{c}{Unlabeled Image \#}                        \\ %\cmidrule(r){2-5}
               & 0      & 1K     & 10K      & 50K                             \\ \midrule
ResNet18       & 56.3   & 62.6   & 66.8     & 71.5                            \\ \midrule
ScatRN18       & 63.1   & 67.2   & 70.8     & 73.2                            \\ \midrule
\textbf{HarmRN18}  & 66.3   & 72.4   & 73.1     & \textbf{78}                   \\ \bottomrule
\end{tabular}
\end{minipage}
\hfill
\begin{minipage}[b]{0.48\linewidth}
\centering
\captionof{table}{The CIFAR10 utility comparison of private classifiers learned on the HarmRN18 features and trained by various DP-training methods (DP-SOTA-2 indicates~\citet{Lou:CVPR2021}).}
\vspace{+0.1in}
\label{t:privacy_cifar10_training}
\begin{tabular}{cccc} \toprule
Public      & Training                 & \multirow{2}{*}{$\epsilon$-DP} & \multirow{2}{*}{Acc (\%)} \\ 
Dataset     & Scheme                   &                                &                           \\ \midrule
            & PATE                     & $16$                           & 70.3                      \\
unlabeled   & DPSGD                    & $1.5$                          & 75.5                      \\
CIFAR100    & DPSGLD                   & $1.5$                          & 74.9                      \\ 
            & \textbf{DPDFA}           & $\mathbf{1.5}$                 & \textbf{78.2}                      \\ \midrule
CIFAR100    & DP-SOTA-2                & $1.5$                          & 71                        \\ \bottomrule
\end{tabular}
\end{minipage}
\vspace{-0.2in}
\end{table}

\textbf{Training a feature extractor}. We applied a series of aggressive data augmentations (e.g., cropping) on a single $600\times225$ image (Appendix~\ref{s:single}) to create a synthetic dataset as the public dataset. More details on our experimental methodology can be found in Section~\ref{s:exp}. To first train a feature extractor on the synthetic dataset, we used three different network architectures including ResNet18~\citet{He:CVPR2016}, ScatRN18, and HarmRN18 to serve as the backbone of the self-supervised SimCLRv2~\cite{Chen:NIPS2020}. ScatRN18 means ResNet18 with ScatterNet handcrafted features, while HarmRN18 indicates ResNet18 where all convolutional filters are replaced by DCT-based HarmonicNet filters. ResNet18 contains four basic blocks connected sequentially. We extracted pretrained features from all four basic blocks, and use ResBlock-$X$ to represent the pretrained features extracted from the block $X$ of ResNet18. We show the CIFAR10 utility comparison of private classifiers learned on the pretrained features produced by different feature extractors under the privacy budget of $(\epsilon=3, \delta=10^{-5})$ in Table~\ref{t:privacy_cifar10_fewimage}. All private classifiers are 1-layer classifiers. Although compared to the last ResBlock, the two middle basic blocks of ResNet18 produce better features, a single image is not enough to make ResNet18 generate high-quality features. Both ScatRN18 and HarmRN18 output better features than ResNet18. The features produced by HarmRN18 (66.3\%) slightly outperform the ScatterNet features (ScatterNet+CNN 66.1\% in Table~\ref{t:privacy_cifar10_noimage})  used directly in the DP training, since DCT is fused with every filter of HarmRN18. We will show the HarmRN18 features can obtain significantly higher utility than the ScatterNet features with the same privacy loss on other complex datasets in Section~\ref{s:no_public_result}.

\subsection{A moderate size public dataset is available}
\label{s:moderate_size}

\textbf{Fine-tuning by public unlabeled images}. To further improve the quality of pretrained features, we randomly selected 1K unlabeled image from a public CIFAR100 dataset to fine-tune the feature extractors, as shown in Table~\ref{t:privacy_cifar10_tuning}, where we still used DPSGD to train the private classifier. Training with 1K unlabeled images improves the utility of the private classifier learned on the features generated by HarmRN18 to $72.4\%$. The entire unlabeled CIFAR100 dataset (50K images) makes the private classifier achieve $78\%$ accuracy on the private CIFAR10 dataset. %Particularly, the first 1K images greatly boost the private classifier utility, while further exponentially enlarging the public dataset size results in only moderate utility improvement.

\textbf{Comparing DP-enabled training frameworks}. We used different DP-enabled training frameworks, i.e., PATE, DPSGD, DPSGLD, and DPDFA, to train a private classifier learned on the HarmRN18 features for a private CIFAR10 dataset. The HarmRN18 feature extractor is trained by a public CIFAR100 dataset. For PATE, teacher and student models are 1-layer linear classifiers, and use the HarmRN18 features. We adopted 1K teacher models, and CIFAR100 as the public dataset. The private classifiers of DPSGD and DPSGLD are 1-layer classifiers, while the private classifier of DPDFA is a 2-layer multilayer perceptron (MLP), where the first layer uses Tanh activations and the second layer uses a Sigmoid activation. We actually tried a 2-layer MLP for all frameworks, but it works better only with DPDFA, as shown in Appendix~\ref{s:2-layer}. The comparison between DP-enabled training frameworks is shown in Table~\ref{t:privacy_cifar10_training}. Compared to DPSGD, the semi-supervised PATE cannot fully take advantage of the pretrained features. The student model has to perform many queries on teacher models to obtain reasonably high utility, so the privacy loss is high ($16$). DPSGLD is similar to DPSGD, but it costs higher privacy overhead in each epoch. As a result, DPSGLD achieves a slightly worse utility than DPSGD under the same privacy budget. Although we used the worst privacy overhead per epoch to estimate the privacy loss of DPDFA, DPDFA still obtains the best utility when $\epsilon=1.5$ among all frameworks. DPDFA makes the private classifier learned on the HarmRN18 features obtain a much better utility than the classifier~\citet{Lou:CVPR2021} learned on the features trained by a labeled CIFAR100 dataset. The comparisons between these DP-enabled training frameworks on more complex private datasets can be found in Section~\ref{s:comp_dp_train}.

\begin{figure*}[t!]
%\vspace{-0.2in}
\centering
\subfigure[Crop (small distance)]{
   \includegraphics[width=1.7in]{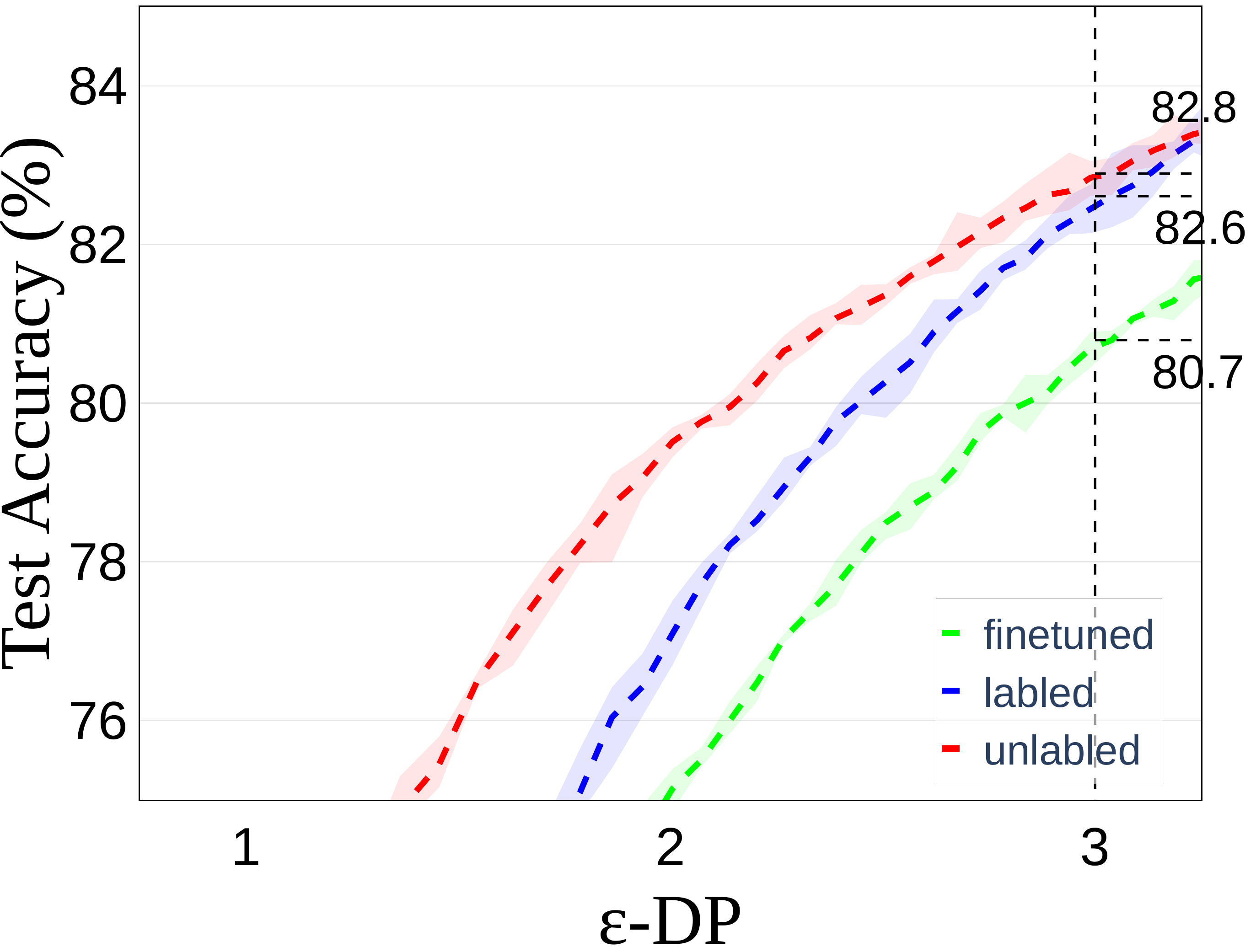}
   \label{f:privacy_miniimagenet_plant}
}
\subfigure[EuroSAT (medium distance)]{
   \includegraphics[width=1.7in]{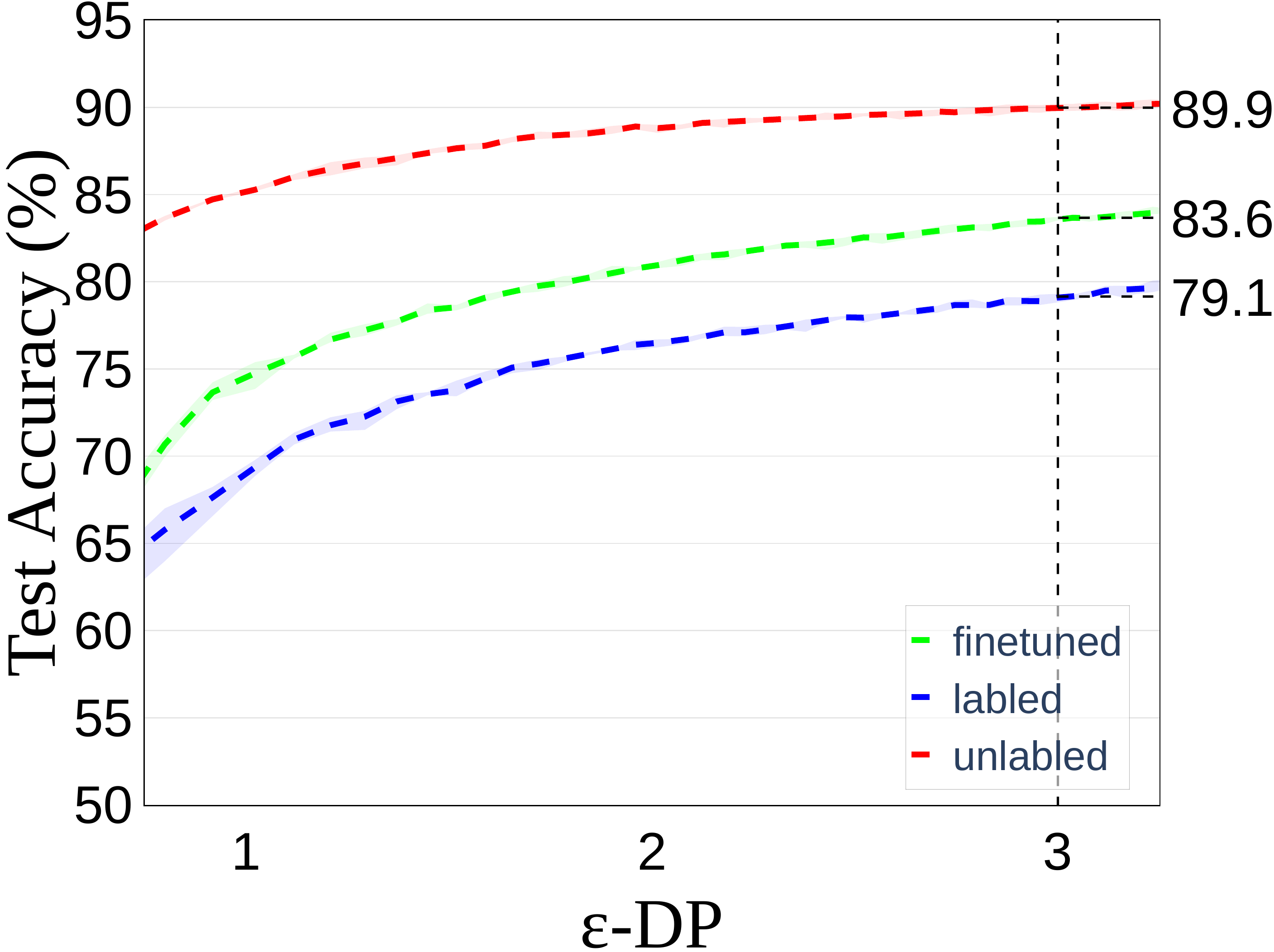}
   \label{f:privacy_miniimagenet_eurosat}
}
\subfigure[ISIC2018 (large distance)]{
   \includegraphics[width=1.7in]{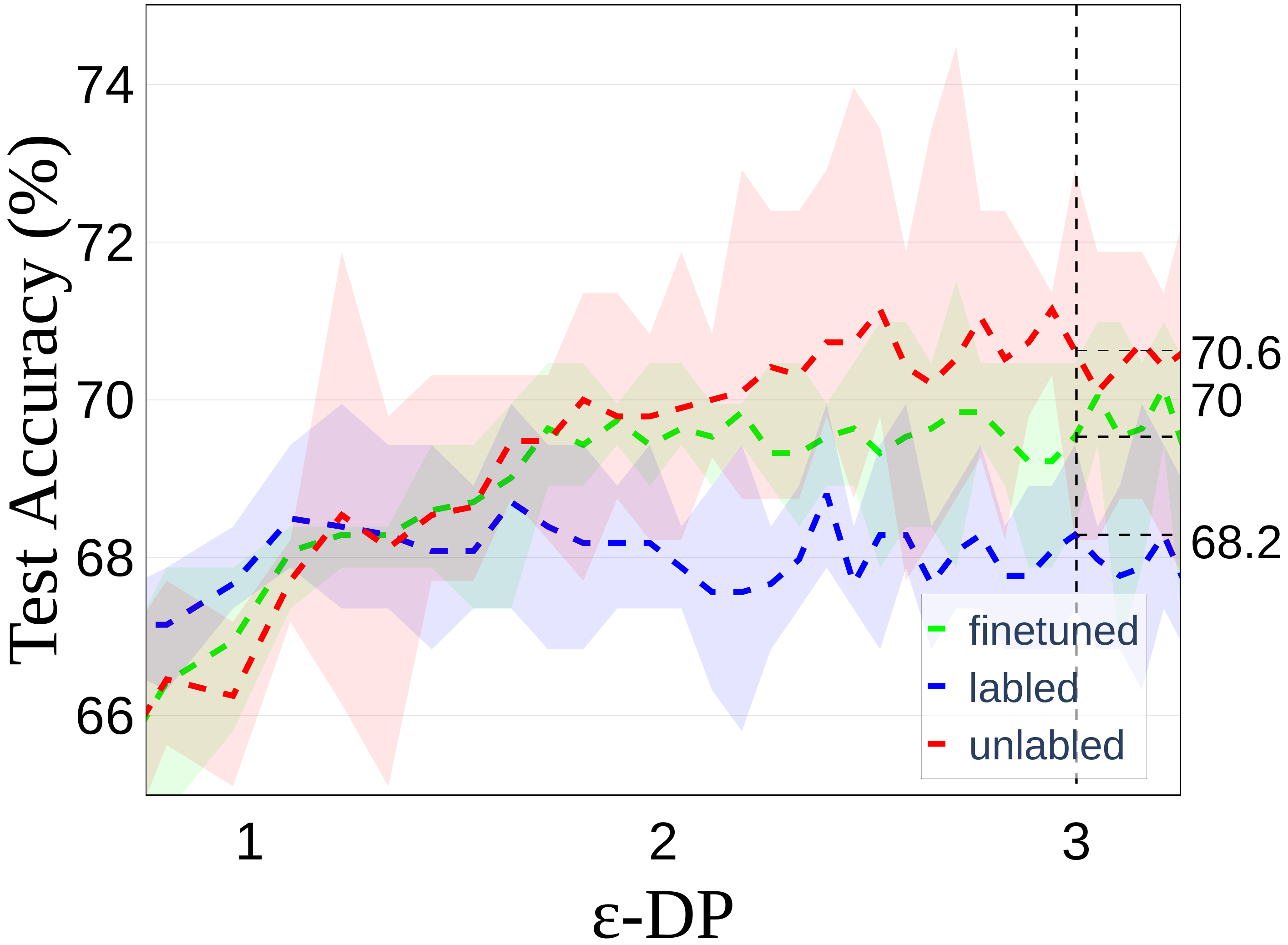}
   \label{f:privacy_miniimagenet_isic2018}
}
\vspace{-0.1in}
\caption{The comparison of features trained by labeled and unlabeled mini-ImageNet on various out-of-domain complex private datasets.}
\label{f:privacy_miniimagenet_training}
\vspace{-0.1in}
\end{figure*}

\textbf{Pretrained features for cross-domain datasets}. To exam the effectiveness of pretrained features on a private dataset from a different domain, we adopted mini-ImageNet~\citet{Vinyals:NIPS2016} as our public dataset to train the pretrained features, and studied the utility-privacy trade-off on out-of-domain private datasets such as CropDiseases~\citet{Mohanty:FIPS2016} (Crop), EuroSAT~\citet{Helber:JSTAEORS2019}, and ISIC2018~\citet{Tschandl:SD2018}. We first trained HarmRN18 by SimCLRv2 on the mini-ImageNet dataset with no label. And then, we used HarmRN18 as the feature extractor to produce features on private Crop, EuroSAT, and ISIC2018 datasets. Finally, we trained a private classifier on each set of features. We trained a ResNet18 feature extractor~\citet{Lou:CVPR2021} by the labeled mini-ImageNet dataset as our baseline. Our baseline is trained using the standard cross-entropy loss with label smoothing. We set the label-smoothing parameter to 0.1. The comparison between our scheme and baseline is shown in Figure~\ref{f:privacy_miniimagenet_training}. Except ISIC2018 under $\epsilon<1.3$, the features generated by SSP enable the private classifier to achieve a better utility than our baseline under the same privacy budget. Particularly, the features produced by SSP on a moderate size of public dataset are already capable enough even for private datasets (e.g., ISIC2018) having a larger learning distance from the public dataset. Compared to supervised pretraining, SSP generates better low- and mid-level features~\citet{Zhao:ICLR2021}, which are more critical to the utility in the DP domain. We also find that further fine-tuning these pretrained features with all labels of the public dataset actually degrades the feature quality on these private datasets. The intra-class invariance~\citet{Zhao:ICLR2021} introduced by the fine-tuning with public dataset labels increases the class misalignment of private datasets.

\begin{figure*}[t!]
\vspace{-0.1in}
\centering
\subfigure[Crop (small distance)]{
   \includegraphics[width=1.7in]{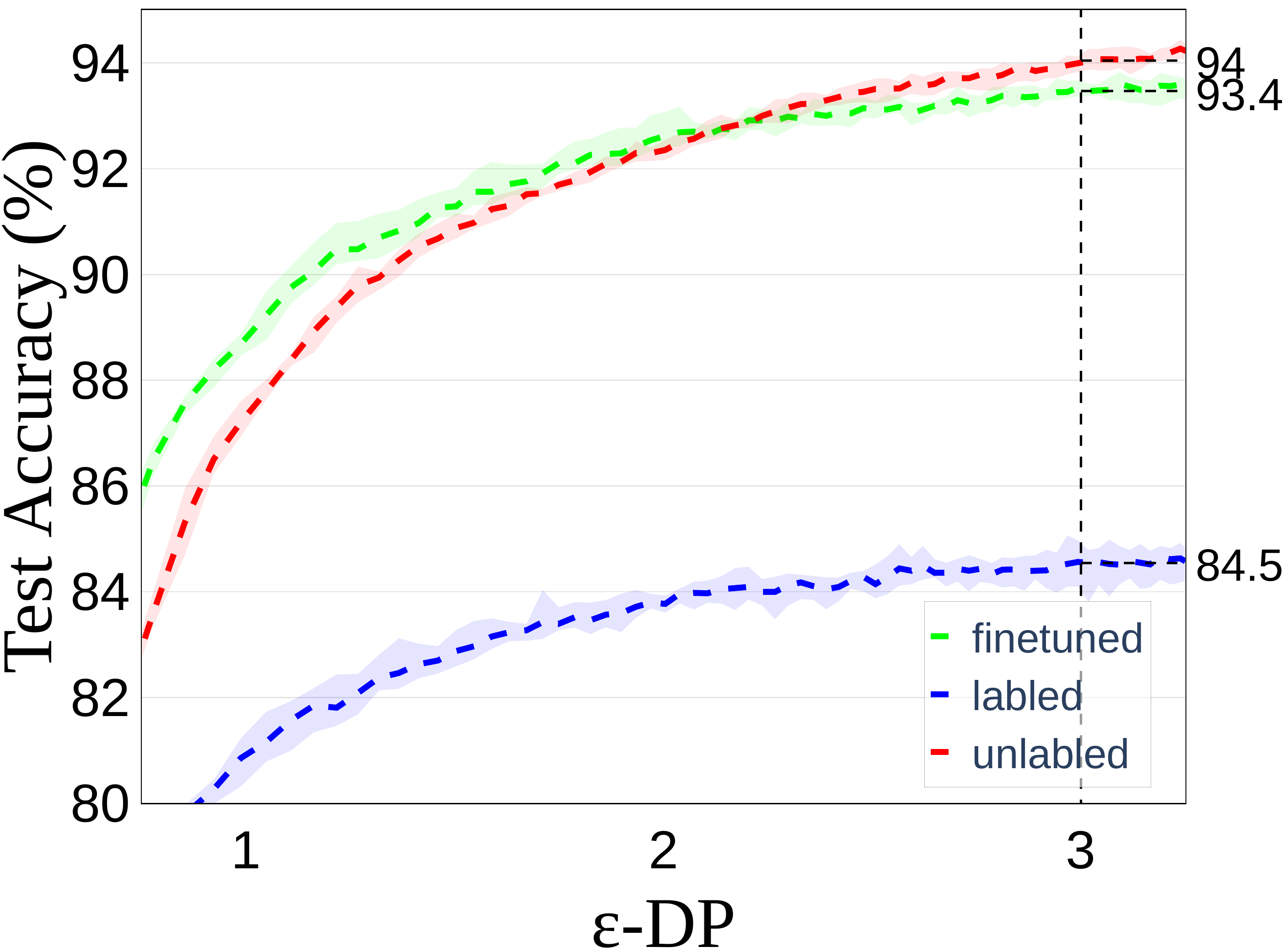}
   \label{f:privacy_imagenet_plant}
}
\subfigure[EuroSAT (medium distance)]{
   \includegraphics[width=1.7in]{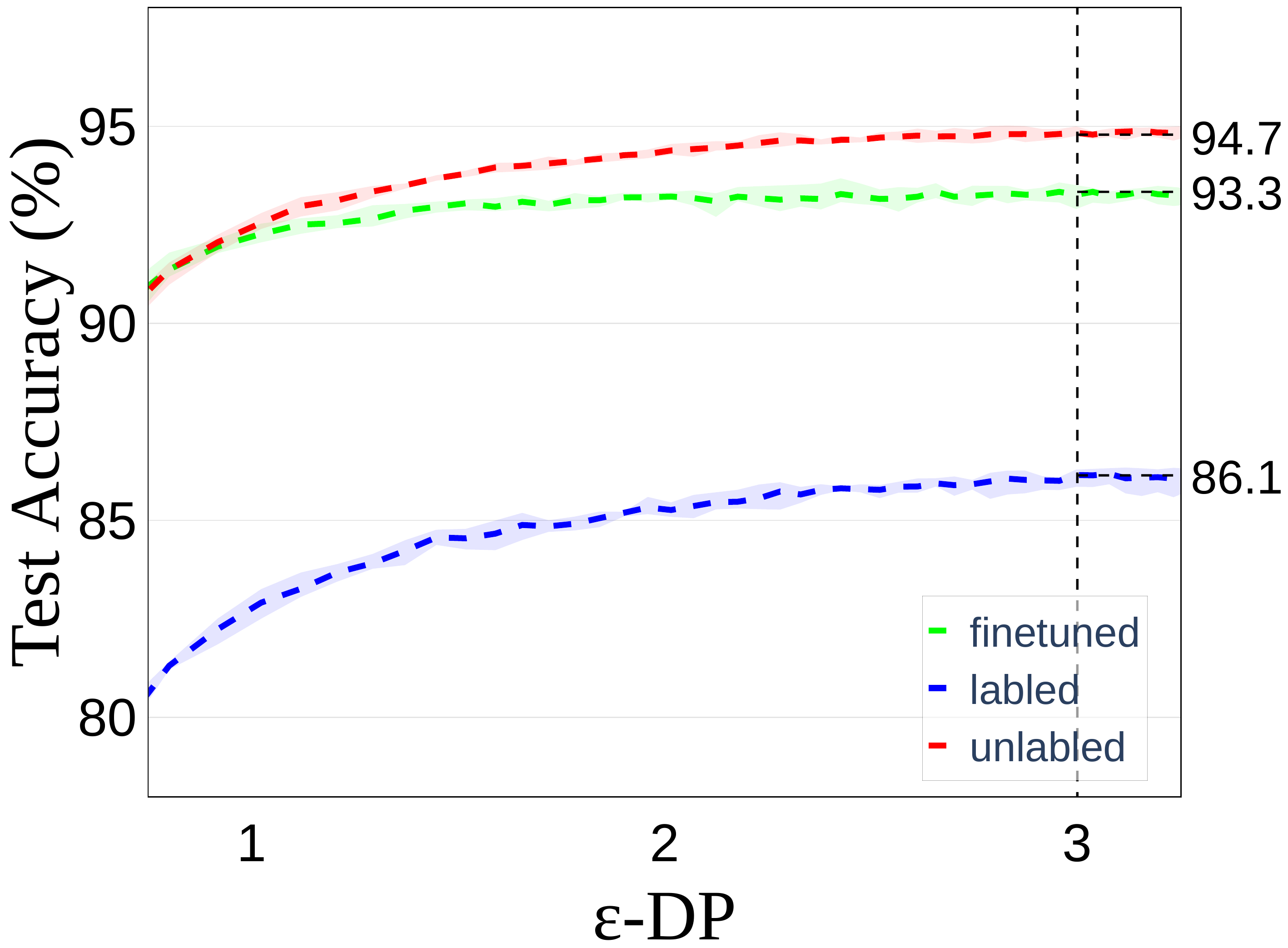}
   \label{f:privacy_imagenet_eurosat}
}
\subfigure[ISIC2018 (large distance)]{
   \includegraphics[width=1.7in]{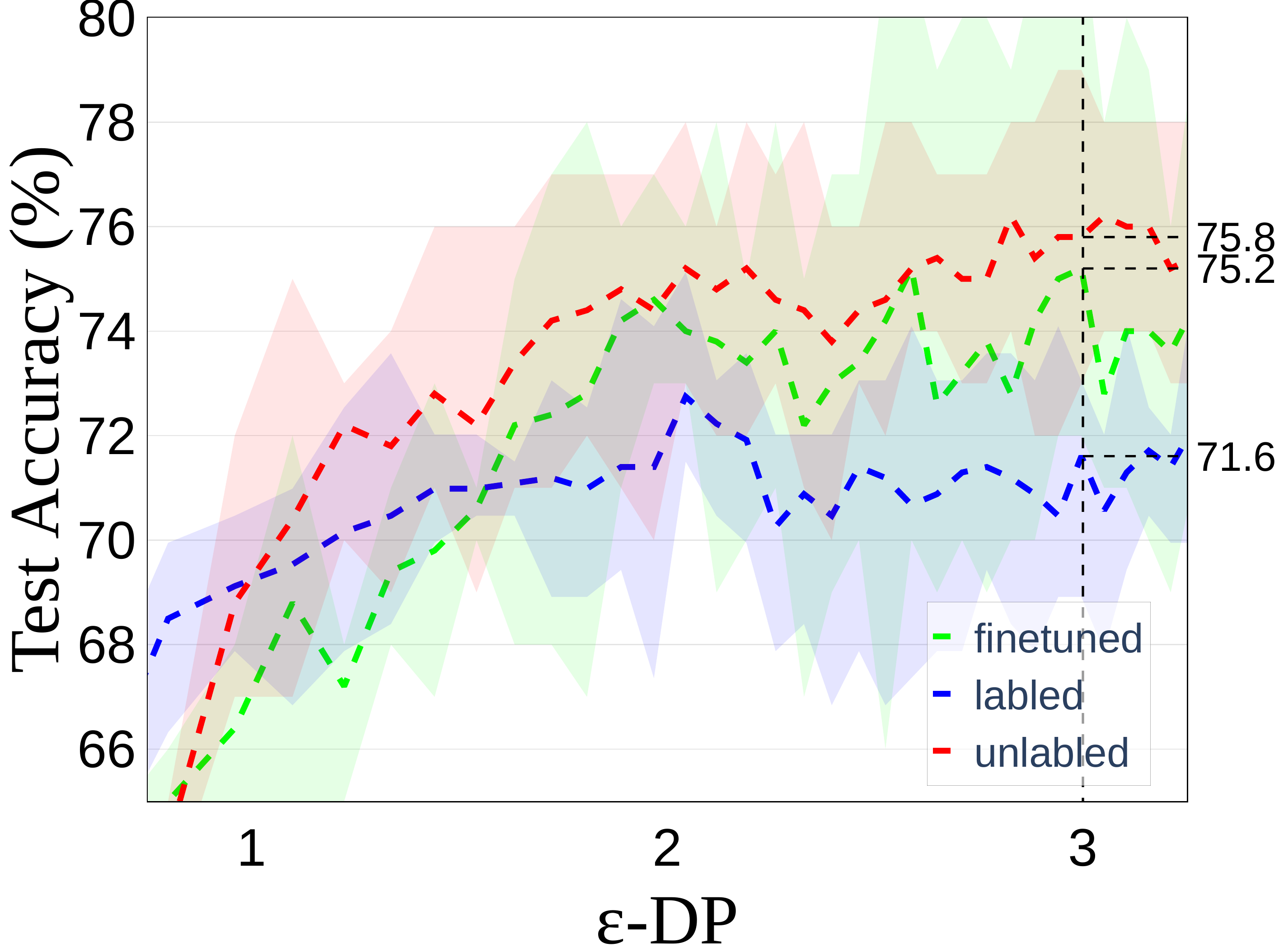}
   \label{f:privacy_imagenet_isic2018}
}
\vspace{-0.1in}
\caption{The comparison of features trained by labeled and unlabeled ImageNet on various out-of-domain complex private datasets.}
\label{f:privacy_imagenet_training}
\vspace{-0.2in}
\end{figure*}

\subsection{A large public dataset is available}

It is easier for a private classifier learned on the features trained by a large public dataset to obtain high utility and small privacy overhead. We adopted ImageNet-1K~\citet{Deng:CVPR2009} as our public dataset to train features with and without labels. And then, we trained a private classifier using these features on various private complex datasets Crop, EuroSAT, or ISIC2018. We used ResNet50 as the backbone of SimCLRv2, and then the feature extractor for private datasets. We also trained another ResNet50 with labels as our baseline. The comparison between our scheme and baseline is shown in Figure~\ref{f:privacy_imagenet_training}. For all three datasets, the features produced by SSP make the private classifier obtain a much higher utility than our supervised-learning baseline under the same privacy budget. A larger unlabeled public dataset greatly improves the quality of features yielded by SSP.

\section{Methods}
\label{s:exp}

\textbf{Public datasets}. We selected the ``Image-B'' in~\citet{Asano:ICLR2020}, which has the size equivalent to $\sim132$ CIFAR images, due to its rich textures and diversity. The image can be viewed in Appendix~\ref{s:single}. We applied a series of aggressive data augmentations including cropping, scaling, rotation, contrast changes, and adding noise on the image, and then created a synthetic dataset having 50K CIFAR/mini-ImageNet-size images. The detailed parameters of these augmentations can be viewed in~\citet{Asano:ICLR2020}. Besides ``Image-B'', we also used CIFAR100, mini-ImageNet~\citet{Vinyals:NIPS2016}, ImageNet~\citet{Deng:CVPR2009}, and PASS~\citet{Asano:PASS2021} as public datasets.

\textbf{Private datasets}. We studied multiple private datasets such as CIFAR10/100~\citet{Krizhevsky:CIFAR10}, CropDiseases~\citet{Mohanty:FPS2016}, EuroSAT~\citet{Helber:JSTAEORS2019}, ISIC2018~\citet{Codella:ISBI2018}, and ImageNet. CropDiseases/EuroSAT/ISIC2018 has a small/medium/large learning distance from the public dataset mini-ImageNet.

\textbf{Feature pretraining}. We adopted SimCLRv2~\cite{Chen:NIPS2020} to train a feature extractor on different public datasets with no label. The backbone of SimCLRv2 uses a ResNet18, ResNet50, or HarmRN18 architecture. The pretraining of each model lasts 200 epochs with a batch size of 64 or 128. After SSP, we used the backbone as an feature extractor.

\textbf{Supervised learning pretraining}. We trained a supervised learning feature extractor on various public datasets using the standard cross-entropy loss with label smoothing. We set the label-smoothing parameter to 0.1.

\textbf{DP-enabled training}. We adopted multiple frameworks including DPSGD~\citet{Abadi:CCS2016}, DPDFA~\citet{Ohana:NIPS2021,Lee:CORR2020}, DPSGLD~\citet{Bu:ICMLTPDPW2021} and PATE~\citet{Papernot:ICLR2018} to train private classifiers. Only DPDFA trains a 2-layer private MLP classifier, while the other DP-enabled training frameworks train a 1-layer linear classifier.

\textbf{Normalization layers}. Models learned on handcrafted features and directly trained on private datasets use only group normalization. All pretrained features are produced with the BN parameters trained by only the public dataset. Results related to BN can be found in Appendix~\ref{s:normalization}.

\textbf{Hyper-parameter search}. Similar to prior work~\citet{Tramer:ICLR2021,Lou:CVPR2021}, we do not count the privacy leakage during hyper-parameter searches on network architectures, optimizers, and hyper-parameters. We target a moderate DP budget of $(\epsilon\leq4.5, \delta=10^{-5})$ for private ImageNet, and a small DP budget of $(\epsilon=0\sim2, \delta=10^{-5})$ for the other private datasets. We fixed the gradient clipping threshold to $C=0.1$ as default, and tried different batch sizes $|B|$ and learning rates $\eta$. The typical batch size we used is 4096, 8192, or 16384. For each set of hyper-parameter, we ran the experiment for five times and report the average values.

\textbf{Library}. We implemented all privacy-related experiments by Opacus v1.1.1~\cite{Yousefpour:NWPML2021}, where native Poisson sampling is supported by BatchMemoryManager. Private models compute per sample gradients, and the DP optimizer does gradient clipping and noise addition.

\section{Results}
\label{s:eval}

To measure a classifier's utility for a range of privacy budgets, we compute the test accuracy and the DP budget $\epsilon$ after each training epoch. For a small DP budget $(\epsilon\leq 2,\delta=10^{-5})$, we studied the features trained with a single image, the public mini-ImageNet dataset, and the public ImageNet dataset. We also compared various DP-enabled training frameworks to train a private classifier with these features on various private datasets. Finally, we studied the utility of a private ImageNet-1K dataset with a moderate privacy budget $(\epsilon\leq4.5, \delta=10^{-5})$.

\begin{table}[t!]
\caption{The utility comparison of features trained by a single image.}
\vspace{+0.1in}
\label{t:privacy_single_result}
\centering
\begin{tabular}{ccccccc} \toprule
Public         & Network             & CIFAR10            & CIFAR100           & Crop               & EuroSAT            & ISIC2018   \\
Dataset        & Architecture        & \multicolumn{5}{c}{Utility ($\%$) under $\epsilon$-DP$=1$ | $2$}                                  \\\midrule
None           & ScatterNet + CNN    & 51.6 | 63.4          & 14.7 | 25.7          & 45.1 | 67.6          & 17.6 | 34.8          & 61.2 | 63.7           \\ \midrule
1 image        & HarmRN18            & \textbf{61.1} | \textbf{64.8} & \textbf{22.3} | \textbf{27.1} & \textbf{70.1} | \textbf{74.3} & \textbf{80.1} | \textbf{84.5} & \textbf{65.1} | \textbf{67.2} \\ \bottomrule
\end{tabular}
\vspace{-0.2in}
\end{table}

%Table 1 shows the maximal test accuracy achieved by a linear ScatterNet model in our hyper-parameter search, averaged over five runs. We also report results with CNNs trained on ScatterNet models, which are described in more detail below. Figure 1 further compares the full privacy-accuracy curves of our ScatterNets and of the CNNs of Papernot et al. (2020b). Linear models with handcrafted features significantly outperform prior results with end-to-end CNNs, for all privacy budgets we consider. Even when prior work reports results for larger budgets, they do not exceed the accuracy of our baseline.

\subsection{No public dataset}
\label{s:no_public_result}

When there is no public dataset, we trained a HarmRN18 network by SimCLRv2 on a single image as our feature extractor. And then, we trained a private 1-layer classifier by DPSGD on various private datasets using features produced by the HarmRN18 feature extractor. As Table~\ref{t:privacy_single_result} shows, the utility of the private classifier is much higher than that of a 5-layer CNN learning directly on the non-learned ScatterNet handcrafted features, when $\epsilon=1$ and $2$. It is difficult to directly learn on even the ScatterNet handcrafted features due to clipped and noisy gradients during DPSGD. The advantage of the features trained by a single image is particularly significant on private datasets like Crop and EuroSAT. This suggests that leveraging features generated by SSP is a scalable solution to differentially private learning even when facing the lack of public datasets.

\begin{table}[t!]
\caption{The utility comparison of features trained by a public mini-ImageNet dataset.}
\vspace{+0.1in}
\label{t:privacy_moderate_result}
\centering
\begin{tabular}{ccccccc} \toprule
Public    & Extractor    & CIFAR10      & CIFAR100     & Crop                & EuroSAT            & ISIC2018      \\
Dataset   & Architecture & \multicolumn{5}{c}{Utility ($\%$) under $\epsilon$-DP$=1$ | $2$}                                                           \\ \midrule
labeled   & ResNet18     & 68.8 | 72.7  & 33.8 | 39.1  & 62.2 | 77            & 69.3 | 76.4          & \textbf{68.4} | 67.8    \\ \midrule
unlabeled & ResNet18     & 69.1 | 73    & \textbf{34.9} | \textbf{40.8}  & 75.2 | 78.1           & 84.9 | 87.6          & 68.1 | \textbf{70.2}            \\ \midrule
unlabeled & HarmRN18     & \textbf{69.5} | \textbf{73.5} & 34.3 | 40.4   & \textbf{76} | \textbf{79.5}    & \textbf{85.2} | \textbf{88.8} & 67.7 | 69.8\\ \bottomrule
\end{tabular}
\vspace{-0.2in}
\end{table}

\begin{table}[t!]
\caption{The utility comparison of features trained by a public ImageNet dataset.}
\vspace{+0.1in}
\label{t:privacy_large_result}
\centering
\begin{tabular}{ccccccc} \toprule
Public         & Extractor           & CIFAR10              & CIFAR100            & Crop               & EuroSAT            & ISIC2018   \\
Dataset        & Architecture        & \multicolumn{5}{c}{Utility ($\%$) under $\epsilon$-DP$=1$ | $2$}                                      \\ \midrule
labeled        & ResNet50            & 90.4 | 91.1            & 61.3 | 65.4           & 81.1 | 83.7          & 82.9 | 85.2          & 69.5 | 72.7      \\ \midrule
unlabeled      & ResNet50            & \textbf{91.6} | \textbf{92.7}   & \textbf{63.3} | \textbf{69.2}  & \textbf{87.7} | \textbf{92.3} & \textbf{92.5} | \textbf{94.3} & \textbf{70.3} | \textbf{75.2} \\ \bottomrule
\end{tabular}
\vspace{-0.1in}
\end{table}

\subsection{A moderate size public dataset}

When a public mini-ImageNet dataset is available, we trained a HarmRN18 network and a ResNet18 network via SimCLRv2 as our feature extractors. We then used their features to train two private 1-layer classifiers by DPSGD on private CIFAR10, CIFAR100, Crop, EuroSAT, and ISIC2018. Compared to our supervised-learning baseline, as Table~\ref{t:privacy_moderate_result} shows, the utility of the private classifiers learned on features produced by SSP is higher than that learned with labels except the ISIC2018 dataset having a large learning distance, when $\epsilon=1$. Although HarmRN18 outperforms ResNet18 in the non-private domain~\citet{Ulicny:BMVC2019}, we do not find there is a significant difference between them in the DP domain. Particularly, the self-supervised pretrained ResNet18 extractor generates slightly better features than HarmRN18 for private CIFAR100 ($\epsilon=1$ and $2$) and ISIC2018 ($\epsilon=2$) datasets. When pretrained on public mini-ImageNet, ResNet18 is strong enough for producing features for various private datasets.

\subsection{A large size public dataset}

A public ImageNet dataset greatly improves the quality of pretrained features for differentially private learning. We used ResNet50 as the backbone of SimCLRv2 and our feature extractor. And then, we trained a private 1-layer classifier by DPSGD on the features of various private datasets. As Table~\ref{t:privacy_large_result} highlights, the utility of the private classifier learned on features produced by SSP is much higher than that learned with labels, when $\epsilon=1$ and $2$. A large public dataset is the key to improving the utility and privacy loss of differentially private learning.

\subsection{Comparing various DP-enabled training frameworks}
\label{s:comp_dp_train}

\begin{figure*}[t!]
%\vspace{-0.1in}
\centering
\subfigure[Crop (small distance)]{
   \includegraphics[width=1.7in]{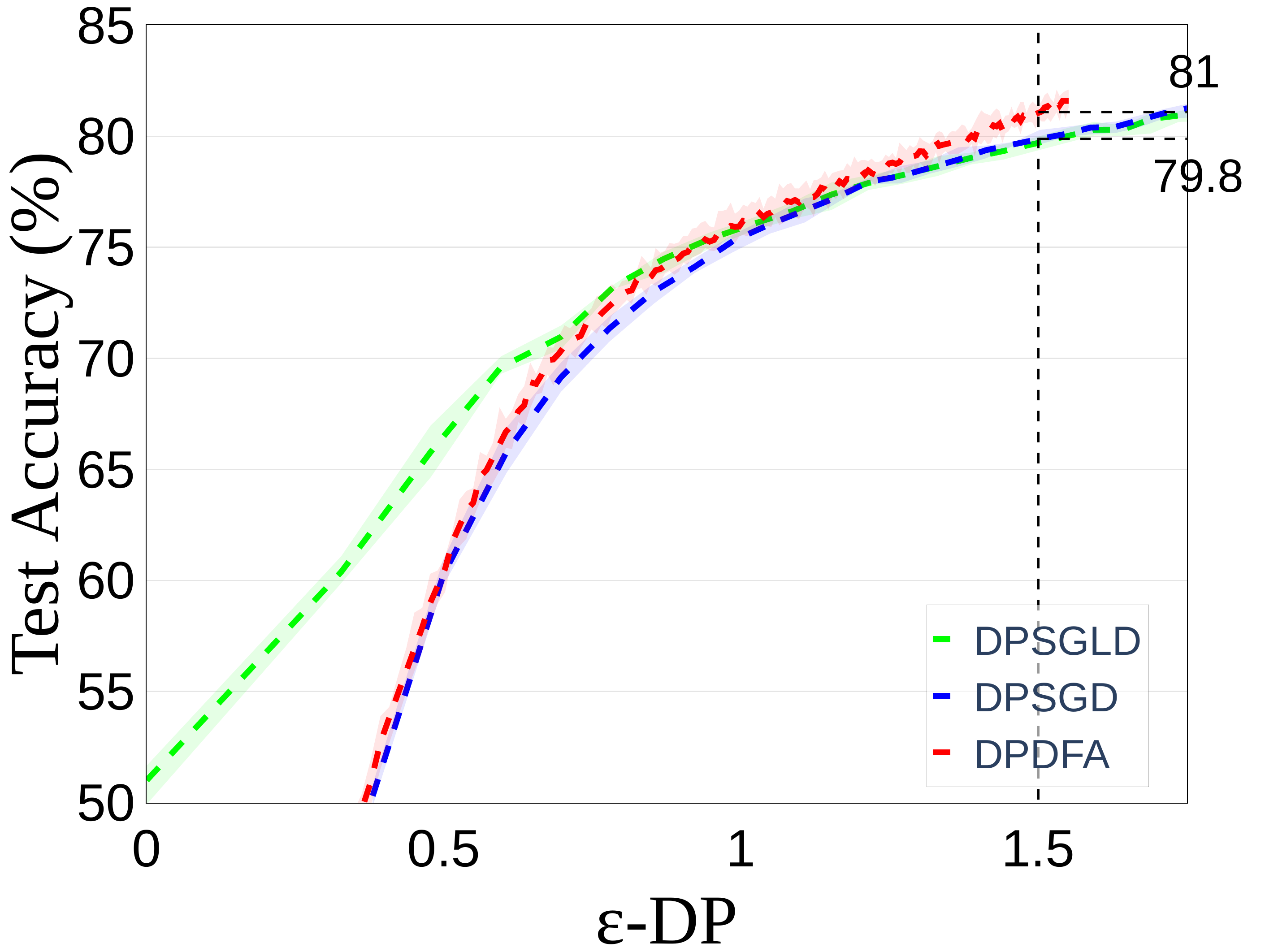}
   \label{f:plants_three_methods_mini}
}
\subfigure[EuroSAT (medium distance)]{
   \includegraphics[width=1.7in]{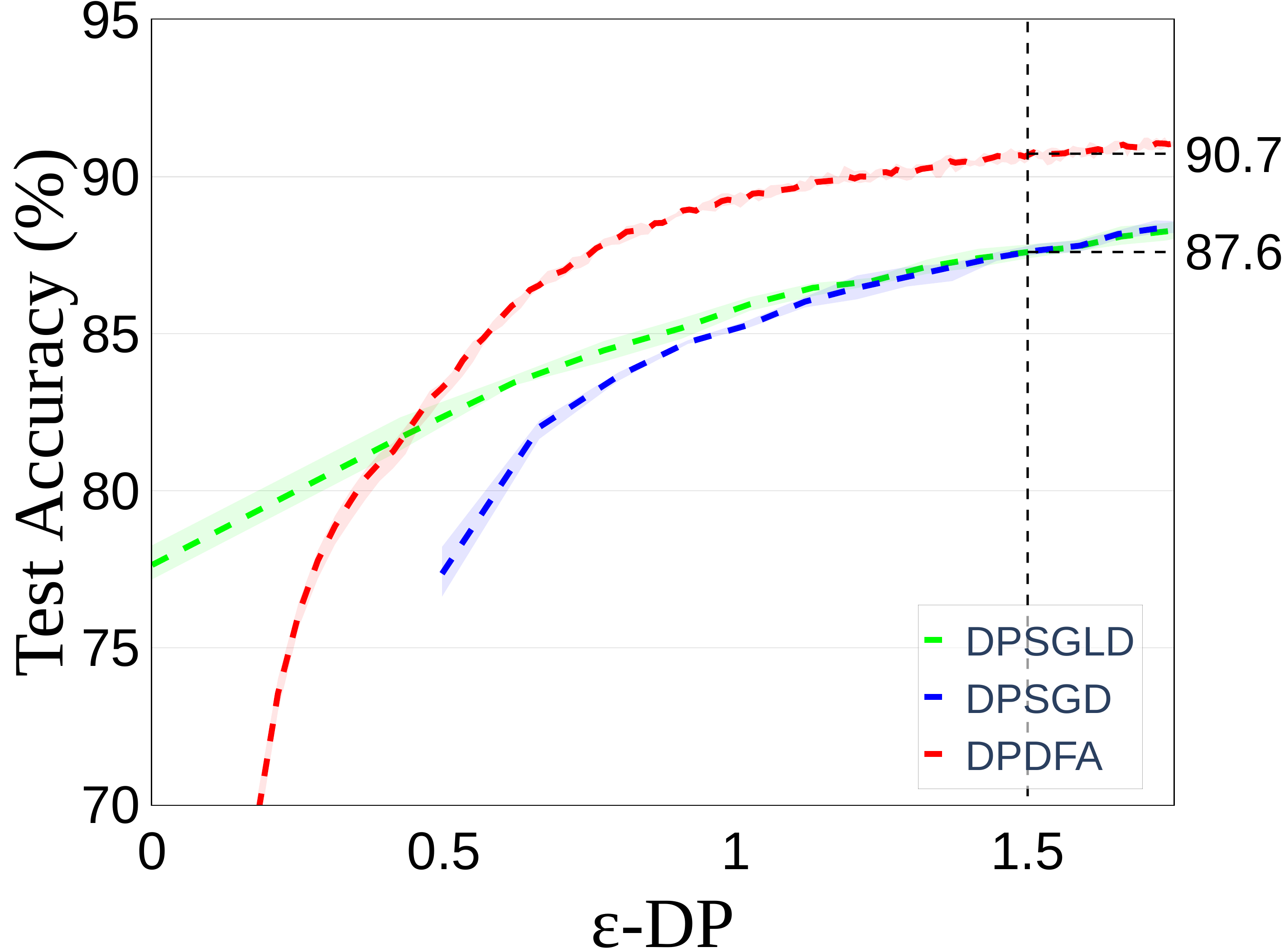}
   \label{f:eurosat_three_methods_mini}
}
\subfigure[ISIC2018 (large distance)]{
   \includegraphics[width=1.7in]{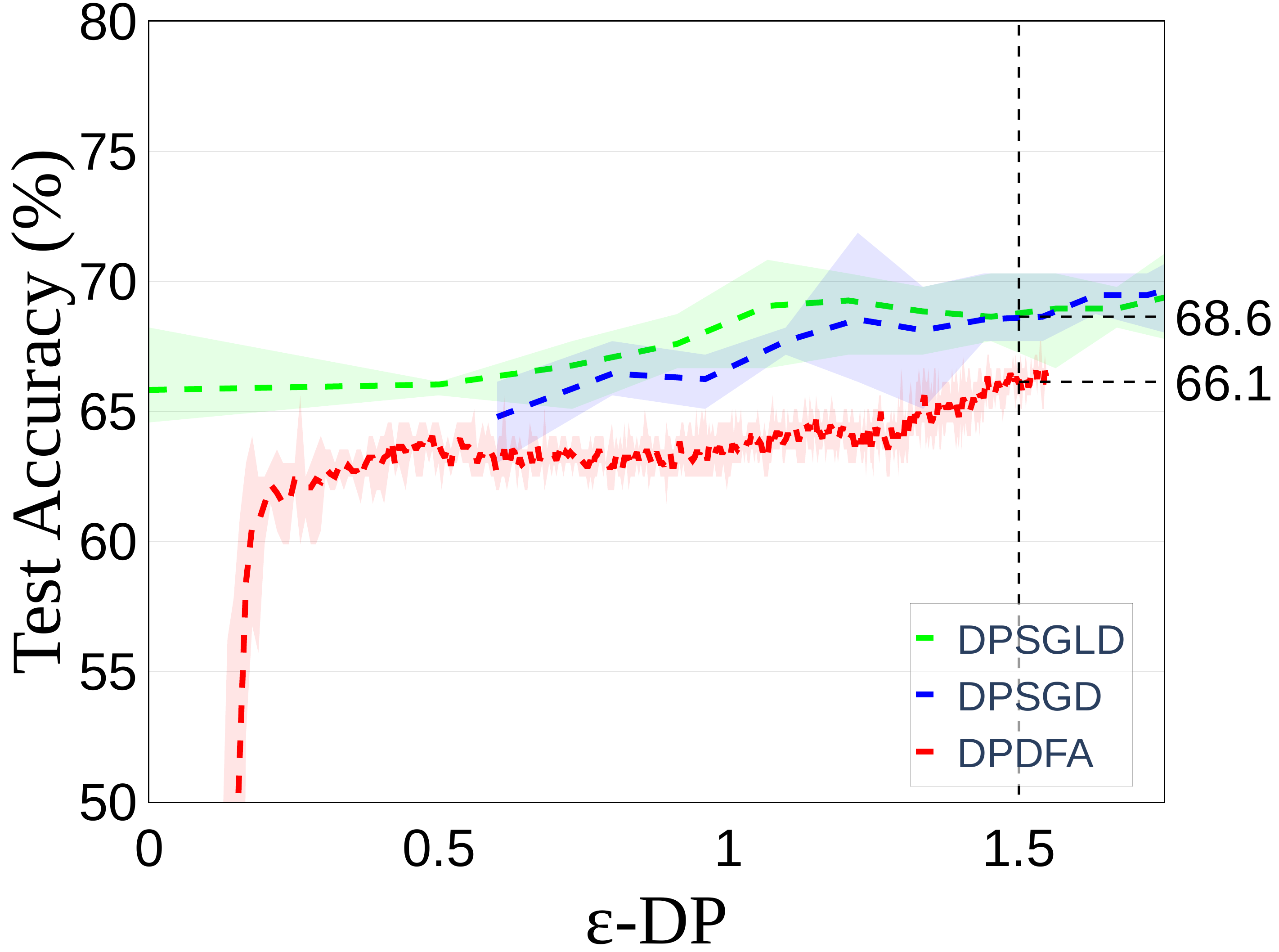}
   \label{f:isic_three_methods_mini}
}
\vspace{-0.1in}
\caption{The comparison of DPSGD, DPSGLD, and DPDFA when training with the features produced by a public mini-ImageNet dataset.}
\label{f:privacy_miniimagenet_framework}
\vspace{-0.2in}
\end{figure*}

\begin{figure*}[t!]
%\vspace{-0.1in}
\centering
\subfigure[Crop (small distance)]{
   \includegraphics[width=1.7in]{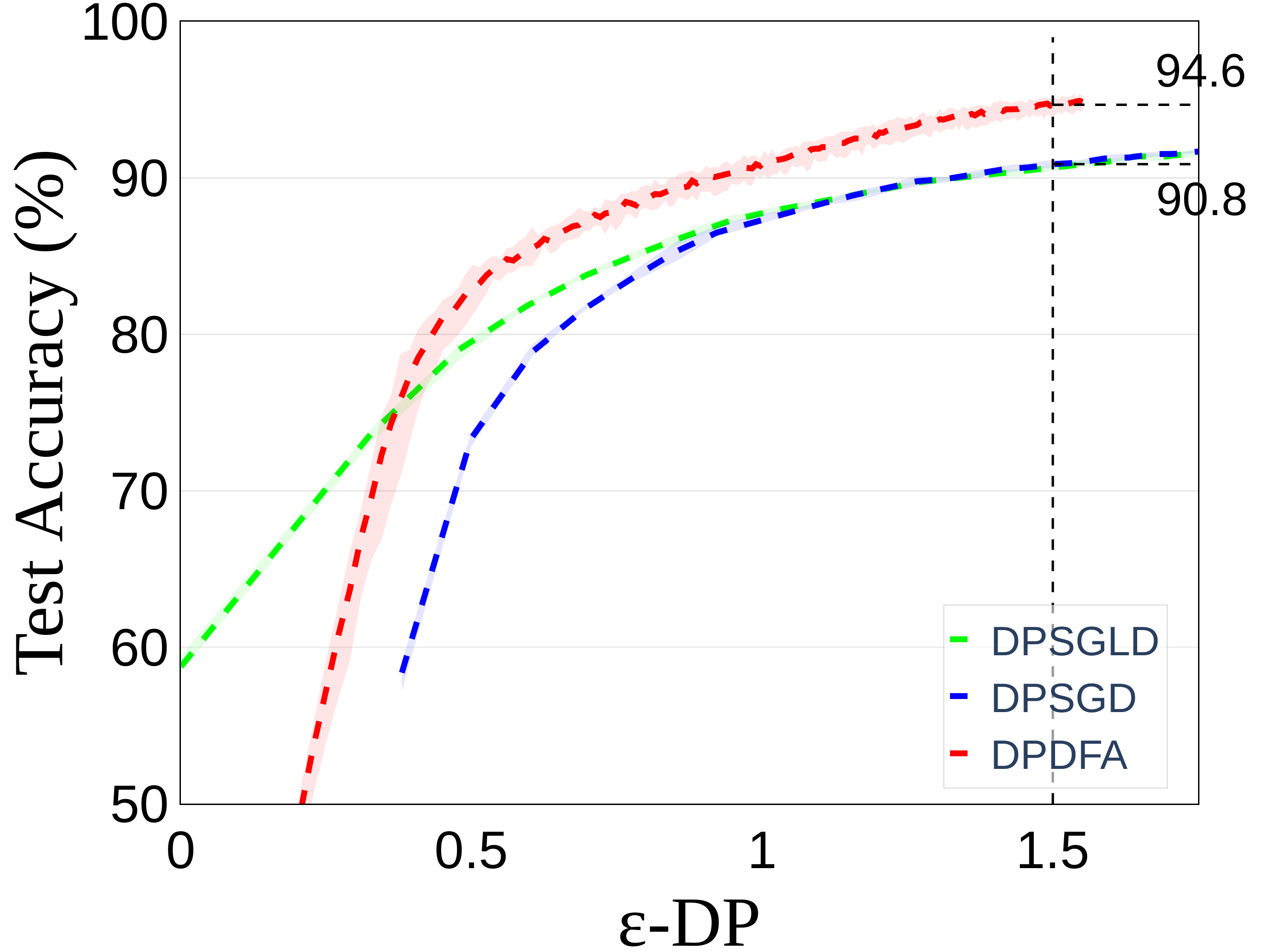}
   \label{f:plants_three_methods_imagenet}
}
\subfigure[EuroSAT (medium distance)]{
   \includegraphics[width=1.7in]{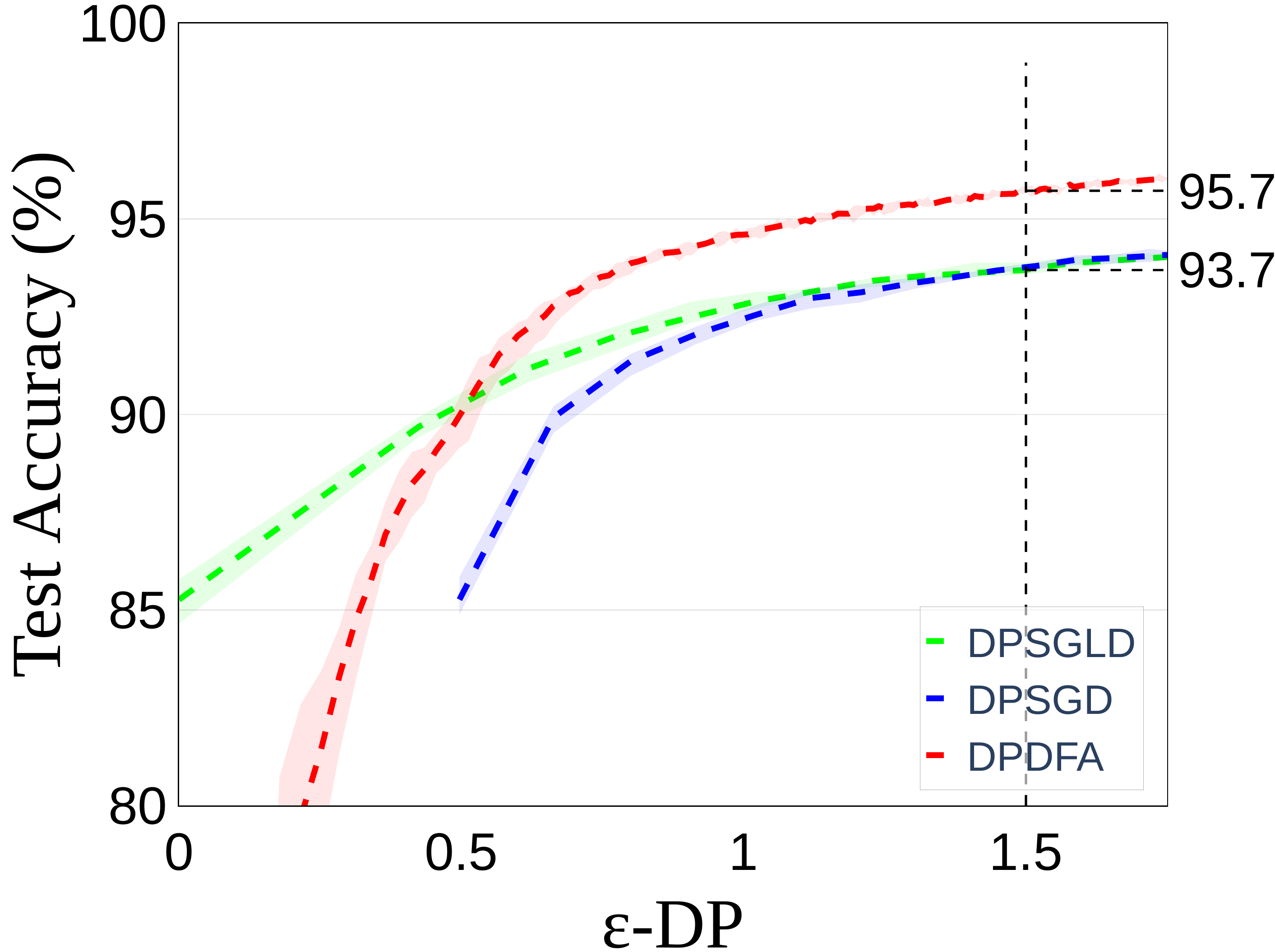}
   \label{f:eurosat_three_methods_imagenet}
}
\subfigure[ISIC2018 (large distance)]{
   \includegraphics[width=1.7in]{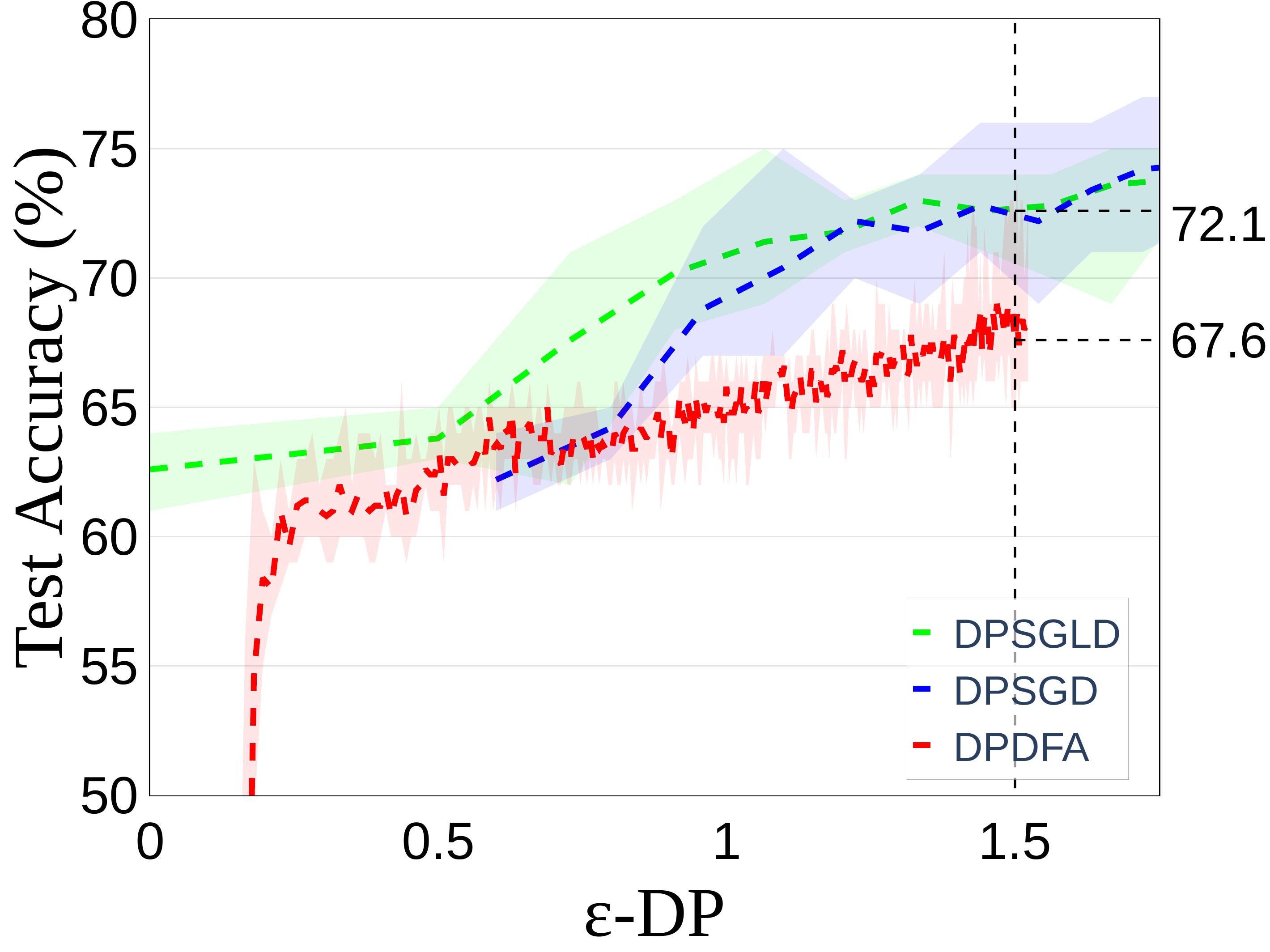}
   \label{f:isic_three_methods_imagenet}
}
\vspace{-0.1in}
\caption{The comparison of DPSGD, DPSGLD, and DPDFA when training with the features produced by a public ImageNet dataset.}
\label{f:privacy_imagenet_framework}
\vspace{-0.2in}
\end{figure*}

We compared DPSGD, DPSGLD, and DPDFA to train a private classifier using the ResNet50 features pretrained on public mini-ImageNet and ImageNet for three private datasets including Crop, EuroSAT, and ISIC2018. Since the utility of PATE is much lower than the other three training frameworks under the same DP budget, we excluded PATE in the comparison. DPSGD is the most widely used DP-enabled training algorithm, so we use it as our baseline. As Figure~\ref{f:privacy_miniimagenet_framework} and~\ref{f:privacy_imagenet_framework} highlight, compared to DPSGD, DPSGLD obtains the same utility, but uses only a smaller privacy budget $\epsilon\leq 1$. However, for a large privacy loss $\epsilon>1$, DPSGD achieves very similar utility to DPSGLD. DPDFA typically can obtain higher utility than the other two DP-enabled training frameworks on the private datasets having a smaller learning distance from the public dataset, i.e., Crop and EuroSAT, under the same privacy budget when $\epsilon>0.5$. However, for private datasets having a large learning distance from the public dataset, i.e., ISIC2018, DPDFA achieves only a lower utility than DPSGD and DPSGLD under the same privacy budget.

\newpage
\subsection{Studying the utility and privacy loss of private ImageNet-1K}
\label{s:utility_privacy_imagenet}

\begin{wrapfigure}[12]{r}{0.32\textwidth}
\vspace{-0.3in}
\centering
\includegraphics[width=1.7in]{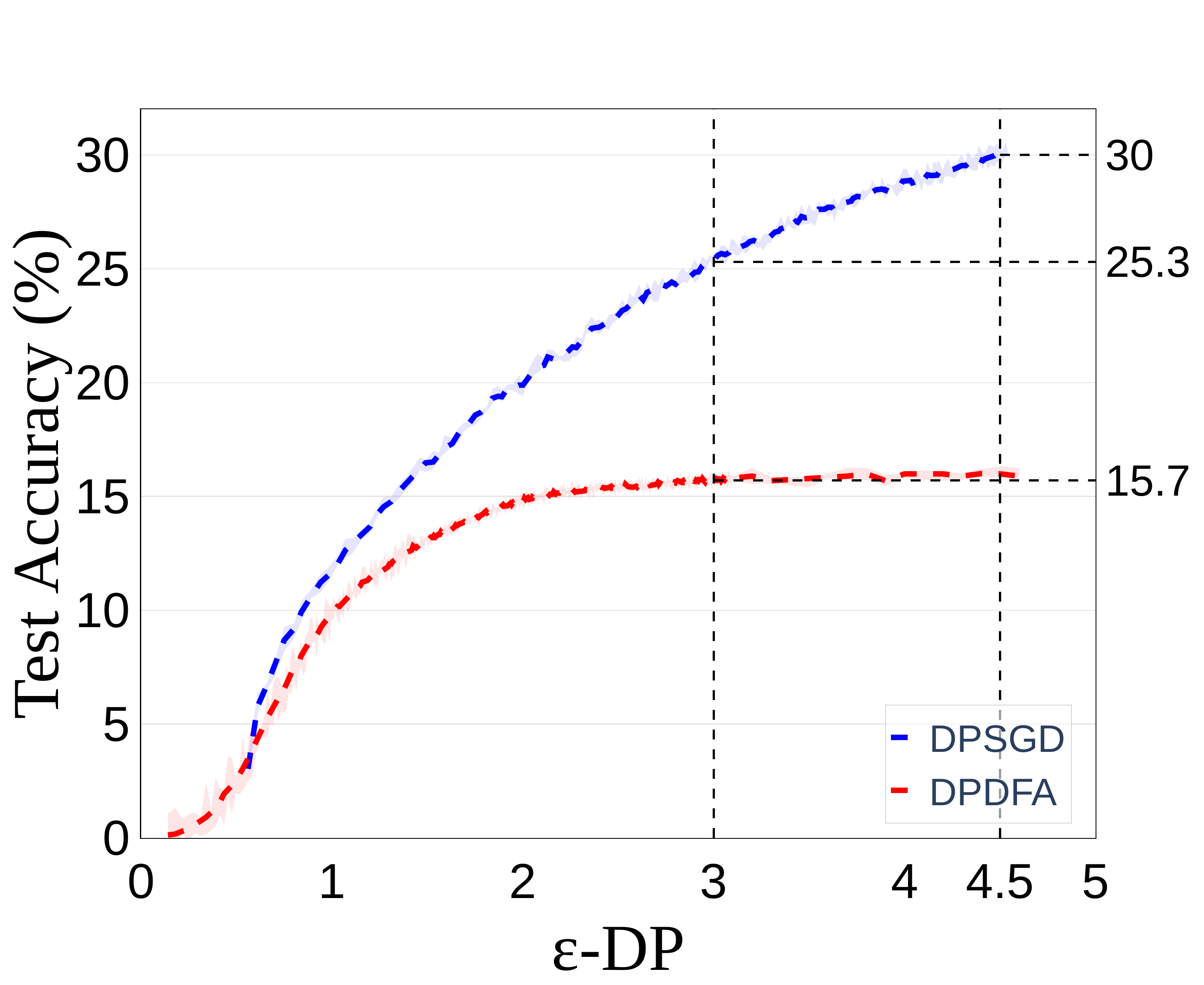}
%\vspace{-0.1in}
\caption{The utility and privacy loss of private ImageNet-1K.}
\label{f:privacy_imagenet_result}
\end{wrapfigure}
We adopted ResNet50 as the backbone of SimCLRv2, and trained it on a public PASS dataset consisting of 1.4 million images with no label. The backbone is used to generate features for private ImageNet. The features can yield 60.8\% top-1 non-private accuracy. We further trained a private classifier on these features in the DP domain. We find that the gradient clipping threshold $C=0.1$ only produces a top-1 utility $8.1\%$ even when $\epsilon=3$. So we enlarged $C$ to 1, and adjusted the other DP parameters. The utility and privacy overhead of the classifier on private ImageNet is shown in Figure~\ref{f:privacy_imagenet_result}. When $\epsilon=3$, the top-1 utility of the classifier becomes 25.3\%. We also compared DPSGD and DPDFA to train our private classifier. Since PASS is still too small to produce strong enough pretained features, DPDFA achieves a lower utility than DPSGD under the same privacy budget. Although recent work~\citet{Kurakin:CORR2022} uses supervised-learning-based features to achieve a top-1 utility 47.9\% on private ImageNet with $(\epsilon=10, \delta=10^{-6})$, its top-1 utility is only 7.6\% when $\epsilon=4.57$. Under a moderate privacy budget ($\epsilon\leq 4.5$), the features generated by SSP greatly outperform prior supervised-learning-based features.

\section{Conclusion and Future Work}
\label{s:con}

We have demonstrated that SSP is a simple yet scalable solution to differentially private learning regardless of the size of available public datasets. The features produced by SSP on a single image, or a moderate/large size public dataset significantly outperform the features trained with labels in the DP domain, let alone the non-learned handcrafted ScatterNet features. Based on the learning distance from the public dataset and the privacy budget, different private datasets may favor distinctive DP-enabled training frameworks to train their private classifiers learned on features produced by SSP. Although we report a non-trivial utility, i.e., 25.3\%, for private ImageNet-1K when $\epsilon=3$, more research efforts are needed to further improve the utility of large-scale private datasets under a moderate privacy budget ($\epsilon\leq3$).

\bibliographystyle{plainnat}
\bibliography{dp}

%%%%%%%%%%%%%%%%%%%%%%%%%%%%%%%%%%%%%%%%%%%%%%%%%%%%%%%%%%%%

\newpage
\appendix

\section{Appendix}

\subsection{R\'enyi DP}
\label{s:renyi}

R\'enyi DP (RDP)~\citet{Mironov:CSF2017} is a generalization of $(\epsilon, \delta)$-DP that uses R\'enyi divergence as a distance metric. The R\'enyi divergence of order $\alpha$ between two distributions $P$ and $Q$ is defined as:
\begin{align*}
D_{\alpha}(P||Q)=\frac{1}{\alpha-1}\log \mathbb{E}_{x\sim P}\left[\left(\frac{P(x)}{Q(x)}\right)^{\alpha-1}\right]
\end{align*}
A model satisfies $(\alpha, \epsilon)$-RDP if 
\begin{align*}
D_{\alpha}(M(D)||M(D'))=\frac{1}{\alpha-1} \log \mathbb{E}_{x\sim M(D)}\left[ \left( \frac{\text{\bf Pr}[M(D)=x]}{\text{\bf Pr}[M(D')=x]} \right)^{\alpha-1} \right]
\end{align*}
Pure $(\epsilon, 0)$-DP is equivalent to $(\infty, \epsilon)$-RDP. And if a model $M$ satisfies $(\alpha, \epsilon)$-RDP, $M$ also satisfies for any $\delta\in(0,1)$. RDP during the training of a model is enforced by two components: per-sample gradients are clipped at a fixed L2 norm threshold $C$, and Gaussian noise of magnitude $\sigma^2 C^2$ is added to the gradient updates for a noise scale parameter $\sigma$.

\begin{table}[htp!]
\caption{The utility and privacy comparison of various schemes using batch normalization on a private CIFAR10 dataset. DP-SOTA-1 means \citet{Tramer:ICLR2021}, and DP-SOTA-2 indicates \citet{Lou:CVPR2021}.}
\vspace{+0.1in}
\label{t:privacy_cifar10_normalization}
\centering
\begin{tabular}{cccccc} \toprule
Public Dataset     & Scheme          & $\epsilon$-DP  & Network     & Training           & Accuracy (\%) \\ \midrule
None               & DP-SOTA-1       & 3              & Scat + CNN  & DPSGD              & 69.3          \\
1 image            & Ours            & 3              & HarmRN18    & DPSGD              & 71.1          \\ \midrule

labeled CIFAR100   & DP-SOTA-2       & 1.5            & ResNet18    & DPSGD              & 81.6          \\ 
unlabeled CIFAR100 & Ours            & 1.5            & ResNet18    & DPSGD              & 80.7          \\  \bottomrule
\end{tabular}
\end{table}

\subsection{Private data normalization}
\label{s:normalization}

Prior work~\citet{Tramer:ICLR2021,Lou:CVPR2021} finds batch normalization (BN) greatly improves convergence on complex private datasets. Private BN~\citet{Tramer:ICLR2021} is proposed to compute private estimates of the per-channel mean and variance of the ScatterNet features. It is difficult to use the same set of hyper-parameters to train BN layers and the other private components in a network. Different values for noise multiplier~\citet{Tramer:ICLR2021} or learning rate~\citet{Lou:CVPR2021} are used to train BN layers. Moreover, most DP training libraries do not support BN layers in a private model yet. We adopt the same method and hyper-parameters to train BN layers as prior work~\citet{Tramer:ICLR2021,Lou:CVPR2021}. The utility and privacy comparison between our schemes and prior work is shown in Table~\ref{t:privacy_cifar10_normalization}. The features trained by HarmRN18 using BN on a single image obtain better utility than the ScatterNet non-learnable handcrafted features. However, the features trained by ResNet18 on a public unlabeled CIFAR100 achieve a slightly worse utility than those trained with labels~\citet{Lou:CVPR2021}. This is because besides BN layers, \citet{Lou:CVPR2021} also fine-tunes convolutional layers in the DP domain.

\subsection{Learning on a single image}
\label{s:single}

\begin{wrapfigure}[9]{r}{0.5\textwidth}
\vspace{-0.1in}
\centering
\includegraphics[angle=270,width=2.7in]{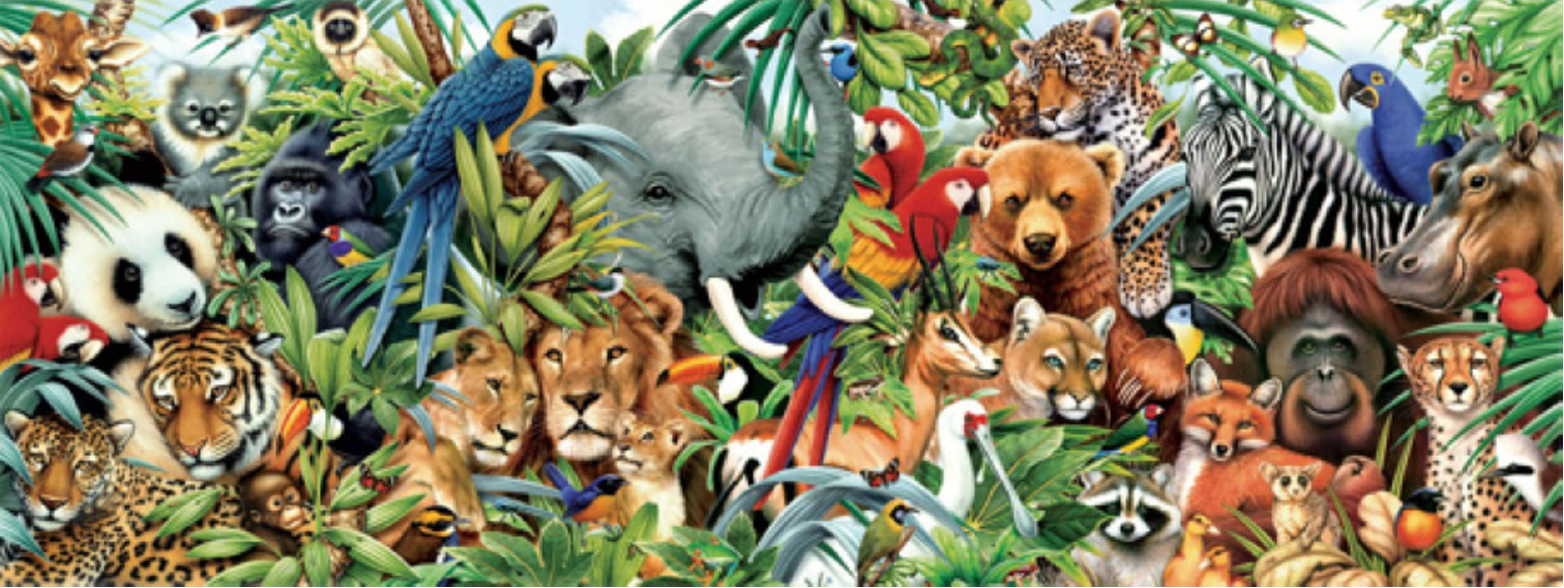}
\caption{Single-image self-supervision.}
\label{f:privacy_cifar10_animals}
\end{wrapfigure}
Recent work~\citet{Asano:ICLR2020} shows the self-supervised learning methods such as BiGAN, RotNet, and DeepCluster can be used to train the first few layers of a deep network model using a single training image, when sufficient data augmentation is used. We select the ``Image-B'', which is shown as Figure~\ref{f:privacy_cifar10_animals}, due to its rich texture and high diversity. The image size is $600\times225$. We also try the ``Image-A'' and the ``Image-C'' from~\citet{Asano:ICLR2020} in our experiments. Among three images, ``Image-B'' achieves the best utility and privacy budget for most private classifiers.

\begin{table}[t!]
\caption{The utility comparison of classifiers consisting of different numbers of layers.}
\vspace{+0.1in}
\label{t:privacy_layernum_result}
\centering
\begin{tabular}{ccccc} \toprule
Public Dataset     & Private Dataset & Arch + Train            & $\epsilon$ & Utility ($\%$)\\ \midrule
                   &                 & 1-layer DPSGD           & $1.5$      & 75.5             \\
                   &                 & 2-layer DPSGD           & $1.5$      & 73.3          \\ 
                   &                 & 1-layer DPSGLD          & $1.5$      & 74.2          \\ 
CIFAR100           & CIFAR10         & 2-layer DPSGLD          & $1.5$      & 72.8              \\ 
								   &                 & 1-layer PATE            & $16$       & 70.2          \\ 
                   &                 & 2-layer PATE            & $16$       & 64.7              \\ 
                   &                 & 2-layer \textbf{DPDFA}  & $1.5$      & \textbf{78.2}              \\ \bottomrule
\end{tabular}
\vspace{-0.1in}
\end{table}

\subsection{Single-layer and 2-layer classifiers}
\label{s:2-layer}

In order to study DPDFA, we need to use a 2-layer linear MLP to serve as our private classifier, where the first layer uses Tanh activations and the second layer uses a Sigmoid activation. A natural question is ``how does DPSGD work with a 2-layer MLP?''. As Table~\ref{t:privacy_layernum_result} shows, we find that unlike DPDFA, all the other DP-enabled training frameworks work better with a single-layer linear classifier. For instance, a 2-layer MLP-based private classifier is always overtopped by a 1-layer classifier when trained by DPSGD. Therefore, in this paper, we always adopt a 2-layer classifier for DPDFA, and a 1-layer classifier for the other DP-enabled training frameworks.

\end{document}